\documentclass[10pt,twocolumn,letterpaper]{article}

\usepackage{cvpr}               % To produce the final version
\usepackage{graphicx}
\usepackage{amsmath}
\usepackage{amssymb}
\usepackage{booktabs}

\usepackage[pagebackref,breaklinks,colorlinks]{hyperref}

\usepackage[capitalize]{cleveref}
\crefname{section}{Sec.}{Secs.}
\Crefname{section}{Section}{Sections}
\Crefname{table}{Table}{Tables}
\crefname{table}{Tab.}{Tabs.}

\def\cvprPaperID{*****} % *** Enter the CVPR Paper ID here
\def\confName{CVPR}
\def\confYear{2022}

\begin{document}

\title{Generative Adversarial Network on Motion-Blur Image Restoration}

\author{Zhengdong Li\\
Department of Electronic and Computer Engineering, HKUST\\
Hong Kong S.A.R, CHINA\\
{\tt\small zlifd@connect.ust.hk}
}

%\and
\maketitle

%%%%%%%%% ABSTRACT
\begin{abstract}
   In everyday life, photographs taken with a camera often suffer from motion blur due to hand vibrations or sudden movements. This phenomenon can significantly detract from the quality of the images captured, making it an interesting challenge to develop a deep learning model that utilizes the principles of adversarial networks to restore clarity to these blurred pixels. In this project, we will focus on leveraging Generative Adversarial Networks (GANs) to effectively deblur images affected by motion blur. A GAN-based Tensorflow model is defined, training and evaluating by GoPro dataset which comprises paired street view images featuring both clear and blurred versions. This adversarial training process between Discriminator and Generator helps to produce increasingly realistic images over time. Peak Signal-to-Noise Ratio (PSNR) and Structural Similarity Index Measure (SSIM) are the two evaluation metrics used to provide quantitative measures of image quality, allowing us to evaluate the effectiveness of the deblurring process. Mean PSNR in 29.1644 and mean SSIM in 0.7459 with average 4.6921 seconds deblurring time are achieved in this project. The blurry pixels are sharper in the output of GAN model shows a good image restoration effect in real world applications. 
\end{abstract}

%%%%%%%%% BODY TEXT
\section{Introduction}
%-------------------------------------------------------------------------
\label{sec:intro}

    The concept of GAN was introduced by Goodfellow \cite{goodfellow2014generative} in 2014. The architecture involves two primary components: the Generator (G) and the Discriminator (D). The Generator is responsible for creating synthetic images that mimic real ones to fool the Discriminator, while the Discriminator's role is to differentiate between real images and the fakes produced by the Generator. Specifically, the Discriminator outputs a value of 1 for real images and 0 for fake images. The objective is to minimize the loss of the Generator while simultaneously maximizing the performance of the Discriminator. This adversarial training process encourages the Generator to produce increasingly realistic images over time. 

\begin{figure}
    \centering
    \includegraphics[width=1\linewidth]{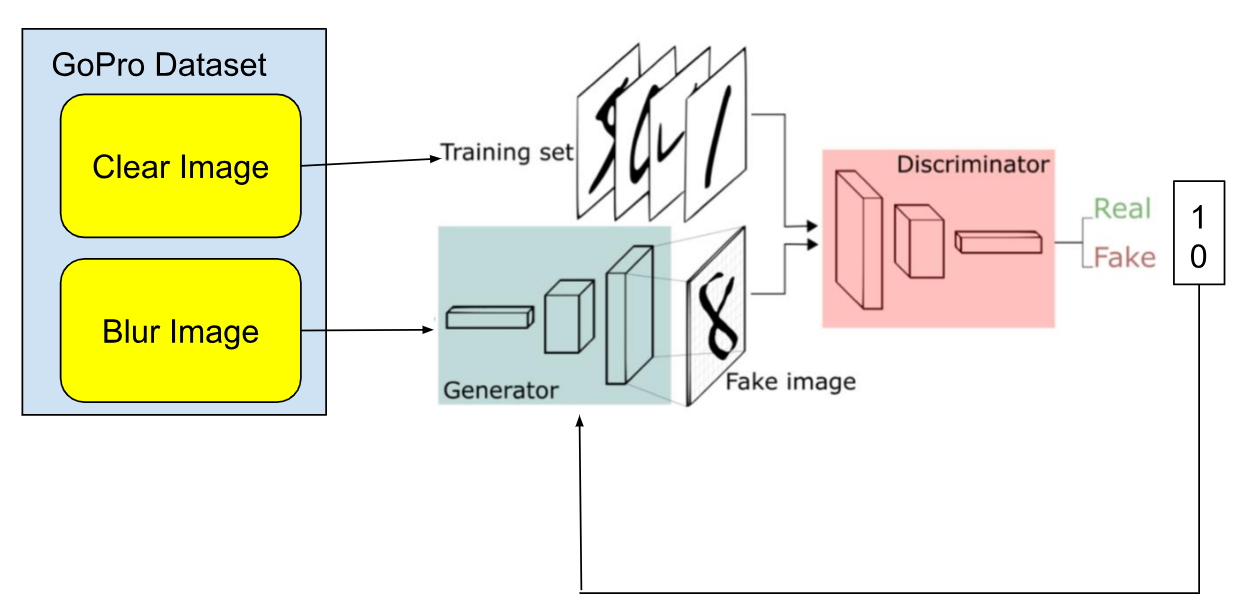}
    \caption{The working flow for Generator and Discriminator in GAN with GoPro dataset}
    \label{fig:GAN}
\end{figure}

\section{Related Work}
%-------------------------------------------------------------------------

    Recently, Kupyn \cite{kupyn2018deblurgan} developed a deblur GAN model in Pytorch framework, but it lacks of details for the architectures of both Generator and Discriminator. The size of the model, like the number of parameters, are also missing. Hence, this project is going to fill in the missing gap of the GAN model in deblurring the motion-blur images. In  our approach, lightweight version of GoPro dataset was used to train the model which was developed using Keras from Tensorflow framework. It provides a pair of both clear and artificially blur images from street views, involving around 500 images with resolution of 1280x720 each in total. We will compare the sharpness of the blurry pixels between input motion-blur images and output deblurred images to see the effectiveness of the pre-trained GAN model. Also, we will compare our performance of the model in terms of different evaluation metrics, like PSNR and SSIM to related works \cite{kupyn2018deblurgan} \cite{kupyn2019deblurgan}. 

\section{Proposed Methods}
%-------------------------------------------------------------------------

    Fig. \ref{fig:GAN} illustrates the working mechanism of image deblurring through  GAN in our proposed method. Unlike feed in the random noise as usual in the GAN model \cite{tran2020gan}, the Generator in our project takes blur images from GoPro dataset as an input directly and generates a synthetic image that closely resembles a real image. Its main objective is to deceive the Discriminator into thinking that the generated image is real. Conversely, the Discriminator receives the synthetic images from Generator and its corresponding real and clear image from GoPro dataset, seeking to accurately identify them. It aims to avoid being misled by the Generator and should ideally classify real images as "True" and fake ones as "False." This feedback loop provides guidance for the Generator to enhance its output, producing more realistic images that can trick the Discriminator over time.

    In this adversarial setup, the Generator (G) and Discriminator (D) have conflicting goals. The Discriminator attempts to maximize its output values, leveraging both the blurry images z and clear images x, while the Generator works to minimize its own output values. This minimax relationship between the two components is key to improving the deblurring effectiveness. The dynamics of this min-max game can be encapsulated in the Eq. \ref{eq:GAN_equation}.

\begin{equation}
  \min_G \max_D V(D, G) = E_x \left[ \log D(x) \right] + E_z \left[ \log \left( 1 - D(G(z)) \right) \right]
  \label{eq:GAN_equation}
\end{equation}

where:

\begin{itemize}
    \item \( G \): The generator function that generates images from blurry images \( z \) from GoPro dataset.
    \item \( D \): The discriminator function that distinguishes between real and generated images.
    \item \( V(D, G) \): The value function representing the adversarial game between the \( G \) and the \( D \).
    \item \( E_x \): The expectation over real images \( x \).
    \item \( D(x) \): The probability that the discriminator correctly identifies a real image.
    \item \( E_z \): The expectation over the blurry images \( z \).
    \item \( G(z) \): The generated image from the blurry images \( z \).
    \item \( D(G(z)) \): The probability that the discriminator identifies the generated image as real.
    \item \( \log D(x) \): The log probability of the discriminator correctly identifying a real image.
    \item \( \log(1 - D(G(z))) \): The log probability of the discriminator incorrectly identifying a generated image as real.
\end{itemize}

    Also, to ensure that the Generator is deblurring the motion-blur images, a perceptual loss function was apply at the output of Generator. The perceptual loss function can be defined in Eq. \ref{eq:loss_equation} as follows:

\begin{equation}
\mathcal{L}_{\text{perceptual}}(y_{\text{true}}, y_{\text{pred}}) = \frac{1}{N} \sum_{i=1}^{N} \left\| \phi(y_{\text{true}}) - \phi(y_{\text{pred}}) \right\|^2
\label{eq:loss_equation}
\end{equation}

where:
\begin{itemize}
    \item \( y_{\text{true}} \) is the ground truth image.
    \item \( y_{\text{pred}} \) is the generated image.
    \item \( \phi \) is the feature extraction model (e.g., VGG16 \cite{simonyan2014very}) applied to the images.
    \item \( N \) is the number of features in the output of layer 'block3\_conv3'.
    \item \( \left\| \cdot \right\| \) denotes the Euclidean norm.
\end{itemize}

\section{Experiments}
%-------------------------------------------------------------------------
\subsection{Training Process}

    The training process starts by segmenting the input image into a 256x256 matrix, which is then fed into the GAN model. The training progresses over the 40 epochs, during which the data is organized with batch size 16. Both the Generator and Discriminator are trained using a constant learning rate of 0.005. At the outset, the Generator creates a synthetic image, while the Discriminator learns to differentiate between this fake image and the real inputs. This iterative process continues to train the entire GAN model. The training persists until all the 500 input images have been utilized. Since our working environment is on MacOS which does not have a CUDA GPU for acceleration, so having longer training time and using smaller dataset become inevitable. The whole training duration typically spans about 4 hours on a 1.4 GHz Quad-Core Intel Core i5 MacOS with single Intel Iris Plus Graphics 645 GPU IDE.

\subsection{GAN Architecture}

    Tab. \ref{tab:GAN_structure} summaries the size of the proposed GAN model in terms of the number of 2D convolutional layers and the number of parameters with 30 layers and 14.5M respectively. Tab. \ref{tab:training_param} also summaries the main training parameters of the GAN model and compared to Kupyn \cite{kupyn2018deblurgan}. 

\begin{figure}
    \centering
    \includegraphics[width=1\linewidth]{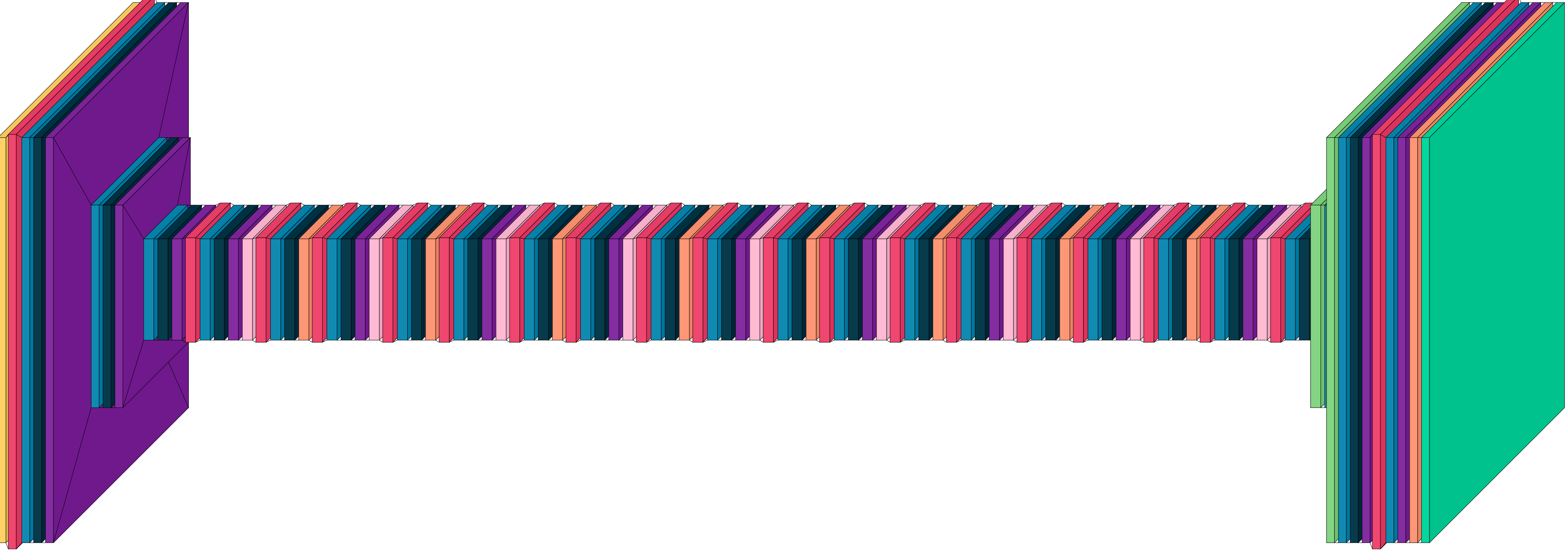}
    \caption{Graphical representation of the architecture of Generator}
    \label{fig:view_model_G}
\end{figure}

    In Generator, 9 ResNet \cite{he2016deep} block are used for sharper image production. In each ResNet, a sequence of Conv2d Layer, following by a Batch Normalization function and ReLU activation function is used. The dropout rate is set to 0.5.  ReLU and Tanh are used as the activation functions with Kernel Size is either 3x3 or 7x7 and Stride Size is either 1x1 or 2x2. Since both input and output of the Generator is an image, so the dimension is 256x256x3 in both head and tail with multiple ResNet blocks in between as shown in Fig. \ref{fig:view_model_G}. 
    
\begin{figure}
    \centering
    \includegraphics[width=1\linewidth]{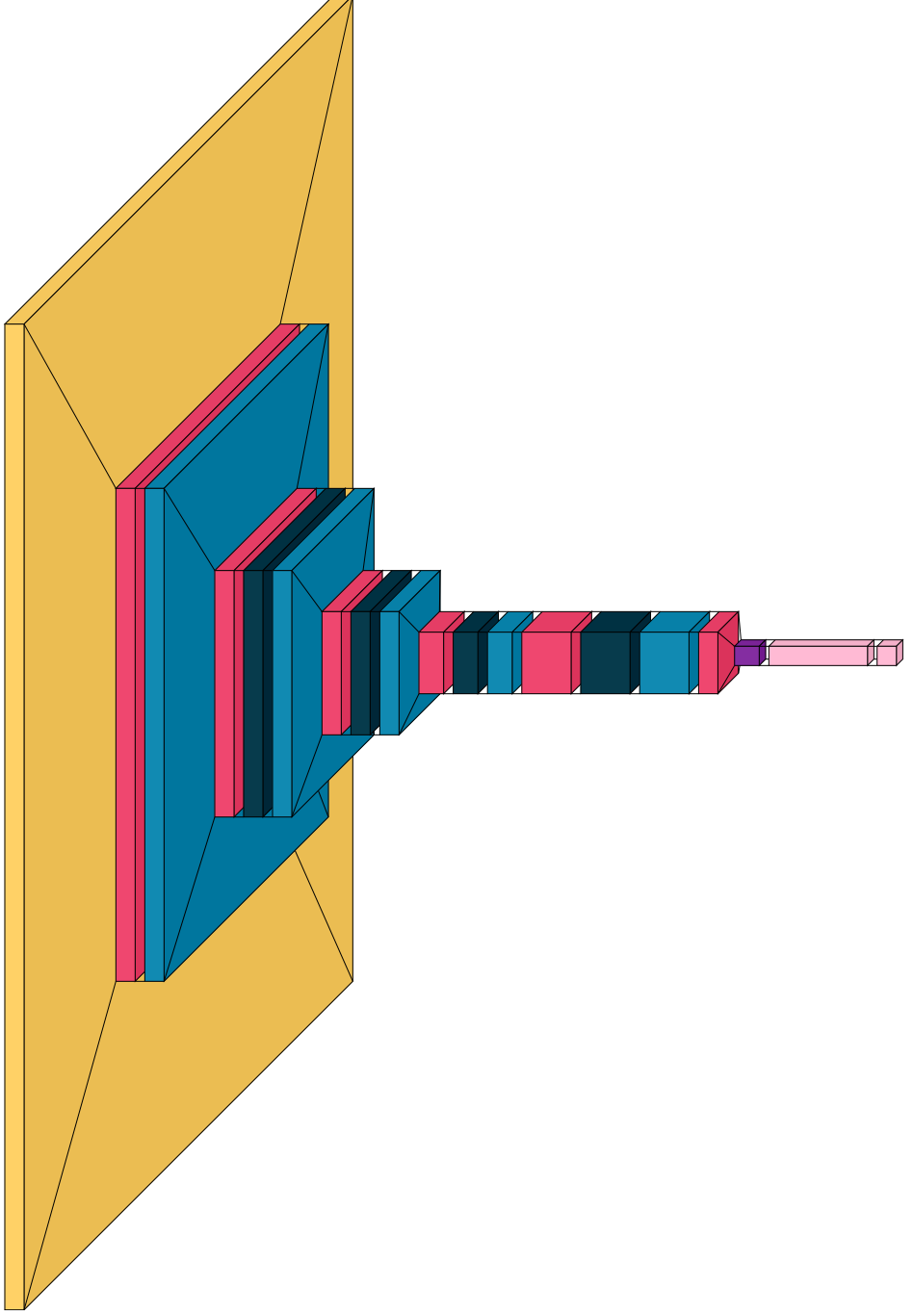}
    \caption{Graphical representation of the architecture of Discriminator}
    \label{fig:view_model_D}
\end{figure}

    In Discriminator, it outputs 0 for fake image and 1 for real image. The activation layers used Sigmoid, LeakReLU and Tanh functions. And the Kernel Size is a constant 4x4 matrix and the Stride Size is the same as Generator. For the Discriminator, its input is also an image while the output is either '1' or '0', hence, the dimension is 256x256x3 in head and 1 in tail as shown in Fig. \ref{fig:view_model_D}.

\begin{table}
    \centering
    \scriptsize
    \begin{tabular}{ccc}
        \toprule
        \textbf{-} & \textbf{this work} & \textbf{Kupyn} \cite{kupyn2018deblurgan}\\
        \midrule
        \textbf{Learning Rate} & 0.005 & 0.0001 \\
        \textbf{Epochs} & 40 & 300 \\
        \textbf{Batch Size} & 16 & 1\\
        \textbf{Dataset} & Light GoPro (~2GB) & Full GoPro (~35GB)\\
        \textbf{Framework} & Keras & Pytorch\\
        \bottomrule
    \end{tabular}
    \caption{Main training parameters in this project compared with Kupyn \cite{kupyn2018deblurgan}}
    \label{tab:training_param}
\end{table}

\begin{table}
    \centering
    \begin{tabular}{ccccc}
         \toprule
         \textbf{Type} & \textbf{Con2d Layers} & \textbf{Params} \\
         \midrule
         Discriminator & 6 & 3.10M \\
         Generator & 24 & 11.40M \\
         \midrule
         \textbf{Total} & 30 & 14.5M \\
         \bottomrule
    \end{tabular}
    \caption{Structure of the GAN model with the number of Convolutional 2D layers and parameters shown. }
    \label{tab:GAN_structure}
\end{table}

\subsection{Results}

    Two metrics, PSNR and SSIM, were utilized to assess the effectiveness of the deblurring process. PSNR (Peak Signal-to-Noise Ratio) quantifies the ratio between the maximum potential power of a signal and the power of noise that degrades its fidelity. It is determined using the logarithm of the ratio of the maximum possible pixel value to the mean squared error (MSE) between the original and processed images. A higher PSNR value signifies a better quality reconstruction.

    On the other hand, SSIM (Structural Similarity Index) evaluates the structural similarity between the original and processed images. It considers three factors: luminance, contrast, and structure. SSIM is computed by analyzing the similarity of these three components across corresponding patches of the original and processed images. The SSIM values range from -1 to 1, where a value of 1 indicates perfect similarity. 

    This project achieves an mean PSNR for 29.1644 and 0.7459 for SSIM as shown in Tab. \ref{tab:PSNR_SSIM_Results}. The average deblurring time per single image is around 4.6921 seconds. Compared to our main reference \cite{kupyn2018deblurgan} and other related works as shown in Tab. \ref{tab:metrics comparison}, we have competitive results for PSNR which is just 0.46 lower than \cite{kupyn2018deblurgan} even we use lightweight version of GoPro dataset. For SSIM, we still have some space to improve, where we are numerically 0.16 to 0.208 behind. One of the reasons might due to the usage of lightweight dataset which is only around 2GB. We believe that the evaluation metrics, like PSNR and SSIM, can be further improved if we a larger dataset, like the original full GoPro dataset which is 35GB in total. 

\begin{figure}
    \centering
    \includegraphics[width=1\linewidth]{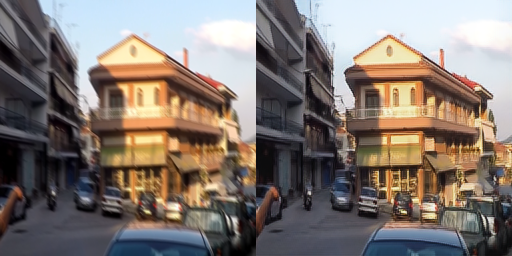}
    \caption{Sample output using blurry image from GoPro dataset in validation part.}
    \label{fig:output01}
\end{figure}

\begin{figure}
    \centering
    \includegraphics[width=1\linewidth]{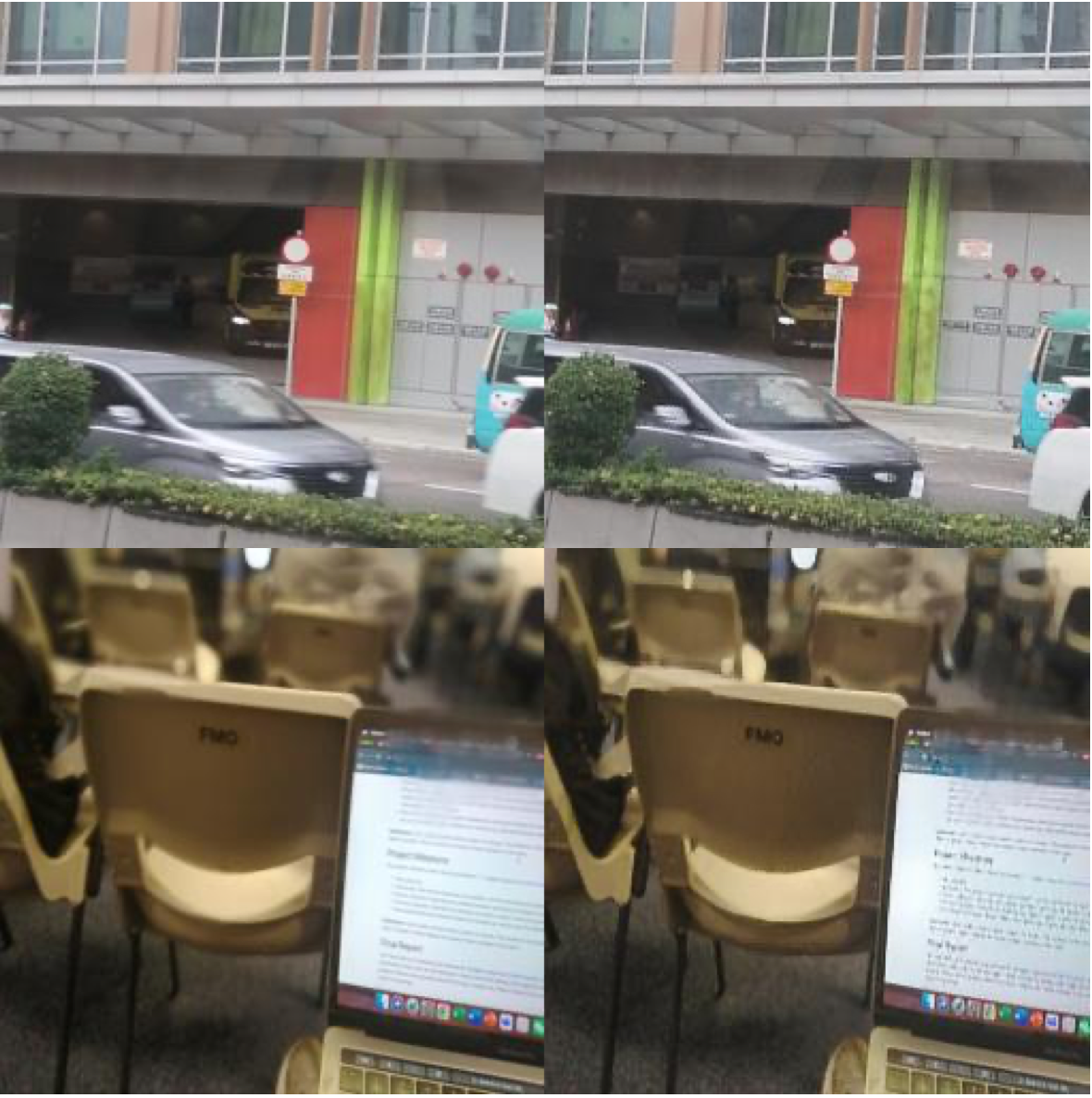}
    \caption{Sample output with self-proposed blurry images.}
    \label{fig:output02}
\end{figure}

\begin{figure}
    \centering
    \includegraphics[width=1\linewidth]{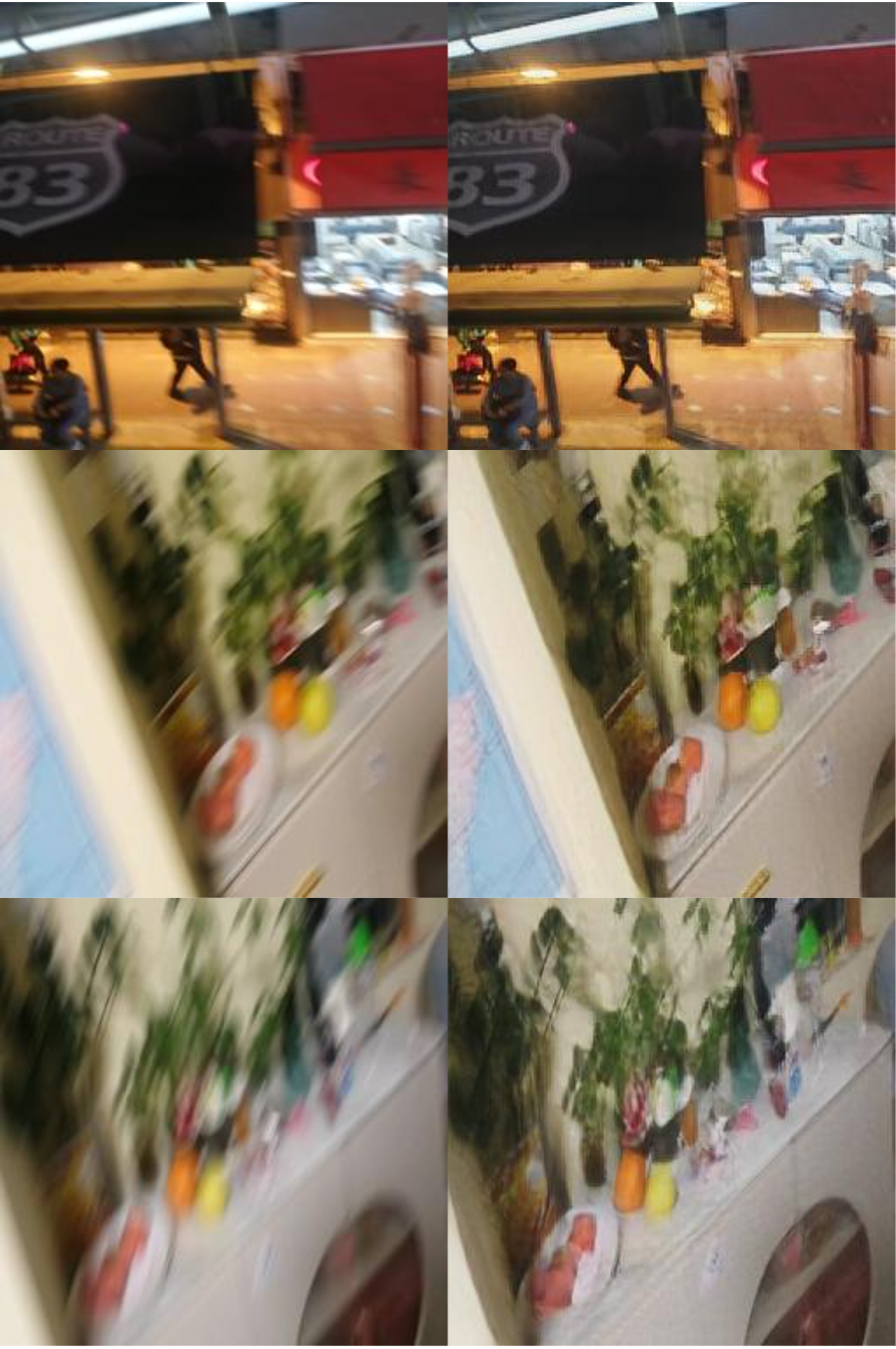}
    \caption{Another sample output with self-proposed blurry images. Since input images have more blurry pixels, so the output images have better deblurred quality. For instance, the edge of the wall is more clear than before in the middle of the figure. }
    \label{fig:output03}
\end{figure}

    For graphical representations, the visualization for deblur performance of the GAN model are shown in Fig. \ref{fig:output01}, Fig. \ref{fig:output02} and Fig. \ref{fig:output03} respectively. Left image is the blurry input and right image is the deblur output. In Fig. \ref{fig:output01}, it is one of the sample image from GoPro dataset in the evaluation part. The training process also use related images, so this type of images has no doubt that having a good deblurred performance. For example, the edges of the whole building are sharper after deblurred by the GAN model. In Fig. \ref{fig:output03}, they are the self-proposed motion-blur images in both indoor and outdoor,  which also yield good deblur quality, and this can be clearly seen in the edges of the wall in the middle part of Fig. \ref{fig:output03}. However,  the deblur performance is not that significant in the self-proposed blurry images when it does not have enough motion-blur pixels from the input images, such as in Fig. \ref{fig:output02}. One reason could be the input image is not fully motion-blurred, which makes it challenging for the Generator to produce a clear image based on the training from the GoPro dataset. Hence, this is also another reason why we do not have a higher SSIM 
    in the previous metrics evaluation part since not all the input images are truly motion-blur.

\begin{table}
    \centering
    \begin{tabular}{cccc}
        \toprule
        Metrics & Highest & Lowest & Mean \\
        \midrule
        PSNR & 30.7574 & 26.4420 & 29.1644 \\
        SSIM & 0.7995 & 0.6499 & 0.7459 \\
        Time & 5.51 & 4.1807 & 4.6921 \\
        \bottomrule
    \end{tabular}
    \caption{Evaluation results for PSNR, SSIM and mean time for deblurring on single image in second on the trained GAN model.}
    \label{tab:PSNR_SSIM_Results}
\end{table}

\begin{table}
    \centering
    \begin{tabular}{ccc}
        \toprule
         - & PSNR $\uparrow$ & SSIM $\uparrow$\\
         \midrule
         this work & 29.16 & 0.75 \\
        \cite{kupyn2018deblurgan} & 28.7 & \textbf{0.958}\\
        \cite{kupyn2019deblurgan} & \textbf{29.55} & 0.934\\
        \cite{nah2017deep} & 28.93 & 0.91\\
         \bottomrule
    \end{tabular}
    \caption{Comparison on PSNR and SSIM in GoPro dataset}
    \label{tab:metrics comparison}
\end{table}

\section{Future Work}
%-------------------------------------------------------------------------
    We aim to improve the deblur performance for the GAN model, yielding higher SSIM and PSNR for evaluation metrics and produce sharper images per pixels for for graphical representations in the future. Hence, the direction of the future work may include but not limited to, (1) construct a deeper neural network architecture, like more convolutional layers and parameters such that the model can better learn the features during training, (2) train with a larger dataset, like \cite{kiani2017need} and \cite{su2017deep} which comprises over 70 real world videos captured in order to generate more training images, (3) work on CUDA GPU environment to shorten the training time, (4) explore different architectures such as U-Net \cite{ronneberger2015u}, which have shown promise in image-to-image translation tasks, capturing more complex features and improving the quality of the generated images, and (5) leverage a pre-trained model as backbone which has already learned useful features from a large dataset can accelerate training and improve performance.

    For our potential application, we aim to utilize a pre-trained GAN model to deblur motion-blurred images captured by smartphones. This innovative approach seeks to address the common issue of poor-quality images that arise from hand vibrations, particularly when capturing fast-moving subjects. By leveraging the capabilities of GANs, we hope to enhance image clarity and restore details that are typically lost during motion blur. However, a significant challenge we face is the difficulty in obtaining authentic motion-blurred images from smartphones. Many modern devices are equipped with advanced optical image stabilization (OIS) \cite{li2012optical} features that effectively mitigate motion blur, resulting in clearer images even in shaky conditions. This technological advancement, while beneficial for everyday photography, poses a hurdle for our research, as it limits the availability of suitable training data for our GAN model. As shown in Fig. \ref{fig:output02}, it is somehow difficult to obtain some motion-blur images. Bad input quality in blurry images results in bad output quality in deblurred images as well. 

    To overcome this obstacle, using existing datasets that contain examples of blurred images or simulating motion blur in controlled environments could be one of the solutions. By addressing these challenges, we aim to develop a robust solution that enhances the quality of smartphone photography, ultimately providing users with clearer, more vibrant images even in less-than-ideal shooting conditions.

\section{Conclusion}
%-------------------------------------------------------------------------

    In this project, we focus on GANs to address the challenge of motion blur in images. By utilizing a GAN-based model implemented in Keras, we focused on training and evaluating our system using the GoPro dataset, which contains paired street view images that include both clear and blurred versions. This dataset provided an ideal foundation for our adversarial training approach, which consists of a Generator and a Discriminator competing with each other. Through this dynamic interaction, the model progressively enhances the quality of the generated images, ultimately leading to more realistic and visually appealing output. Throughout our experiments, we achieved a notable mean PSNR of 29.1644 and a mean SSIM of 0.7459, indicating a significant enhancement in image clarity and structural fidelity compared to the original blurred images. Additionally, the average deblurring time of 4.6921 seconds demonstrates the efficiency of our approach in real-world applications. The results obtained from our GAN model highlight the effectiveness of using adversarial training for image restoration. The output images exhibit sharper details and improved clarity, showcasing the potential of GANs in tackling real-world challenges associated with motion blur. This work not only underscores the advancements in image processing techniques but also opens avenues for further research and development in the field of computer vision, where GANs can be applied to various image enhancement tasks. Overall, our findings contribute to the growing body of knowledge surrounding generative models and their practical applications in restoring image quality.

%%%%%%%%% REFERENCES
{\small
\bibliographystyle{ieee_fullname}
\bibliography{egbib}
}

\end{document}